\documentclass[11pt,a4paper]{article}
  \usepackage[hyperref]{emnlp2018}
  \usepackage[hyphenbreaks]{breakurl}
  \usepackage{times}
  \usepackage[multiple]{footmisc}
  \usepackage{amsmath,amsfonts,amssymb}  
  \usepackage{latexsym}
  \usepackage{hhline}
  \usepackage{multirow}
  \usepackage{url}
  \usepackage{placeins}
  \usepackage{algorithm}
  \usepackage{algorithmic}
  \usepackage{siunitx}
  \usepackage{enumitem}

  \aclfinalcopy 
  
  \DeclareMathOperator*{\argmax}{arg\,max}
  \title{Guided Neural Language Generation for Abstractive Summarization using Abstract Meaning Representation}
  
  \author{Hardy \\
  The University of Sheffield \\
    {\tt hhardy2@sheffield.ac.uk} \\
    \And
    Andreas Vlachos \\
    The University of Sheffield \\
    {\tt a.vlachos@sheffield.ac.uk} \\}
  
  \date{}
  
\begin{document}
  \maketitle
  \begin{abstract}
  Recent work on abstractive summarization has made progress with neural encoder-decoder architectures. However, such models are often challenged due to their lack of explicit semantic modeling of the source document and its summary.
  In this paper, we extend previous work on abstractive summarization using Abstract Meaning Representation (AMR) with a neural language generation stage which we guide using the source document. 
  We demonstrate that this guidance improves summarization results by 7.4 and 10.5 points in ROUGE-2 using gold standard AMR parses and parses obtained from an off-the-shelf parser respectively. We also find that the summarization performance using the latter is 2 ROUGE-2 points higher than that of a well-established neural encoder-decoder approach trained on a larger dataset. Code is available at \url{https://github.com/sheffieldnlp/AMR2Text-summ}
  \end{abstract}
  
  \section{Introduction}
  Abstractive summarization is the task of automatically producing the summary of a source document through the process of paraphrasing, aggregating and/or compressing information. Recent work in abstractive summarization has made progress with neural encoder-decoder architectures \cite{See2017,Chopra2016, Rush2015}. However, these models are often challenged when they are required to combine semantic information in order to generate a longer summary \citep{wiseman-shieber-rush:2017:EMNLP2017}. To address this shortcoming, several works have explored the use of Abstract Meaning Representation \cite[AMR]{Banarescu2013a}. These were motivated by AMR's capability to capture the predicate-argument structure which can be utilized in information aggregation during summarization.
  
  However, the use of AMR also has its own shortcomings. While AMR is suitable for information aggregation, it ignores aspects of language such as tense, grammatical number, etc., which are important for the natural language generation (NLG) stage that normally occurs in the end of the summarization process. Due to the lack of such information, approaches for NLG from AMR typically infer it from regularities in the training data \cite{Pourdamghani2016,Konstas2017,Song2016a,Flanigan2016}, which however is not suitable in the context of summarization. Consequently, the main previous work on AMR-based abstractive summarization \citep{Liu2015a} only generated bag-of-words from the summary AMR graph.
  
  In this paper, we propose an approach to guide the NLG stage in AMR-based abstractive summarization  using information from the source document. Our objective is twofold: (1) to retrieve the information missing from AMR but needed for NLG and (2) improve the quality of the summary. We achieve this in a two-stages process: (1) estimating the probability distribution of the side information, and (2) using it to guide a \citet{Luong2015}'s seq2seq model for NLG.
      
  Our approach is evaluated using the Proxy Report section from the AMR dataset \cite[LDC2017T10]{Knight2017} which contains manually annotated document and summary AMR graphs. Using our proposed guided AMR-to-text NLG, we improve summarization results using both gold standard AMR parses and parses obtained using the RIGA \cite{Barzdins2016} parser by 7.4 and 10.5 ROUGE-2 points respectively. Our model also outperforms a strong baseline seq2seq model \cite{See2017} for summarization by 2 ROUGE-2 points.
  
  \section{Related Work}
  
  \paragraph{Abstractive Summarization using AMR:} In \citet{Liu2015a} work, the source document's sentences were parsed into AMR graphs, which were then combined through merging, collapsing and graph expansion into a single AMR graph representing the source document. Following this, a summary AMR graph was extracted, from which a bag of concept words was obtained without attempting to form fluent text. \citet{Vilca2017} performed a summary AMR graph extraction augmented with discourse-level information and the PageRank \cite{Page1998} algorithm. For text generation, \citet{Vilca2017} used a rule-based syntactic realizer \citep{Gatt:2009:SRE:1610195.1610208} which requires substantial human input to perform adequately. 
  
  \paragraph{Seq2seq using Side Information:} In Neural Machine Translation (NMT) field, recent work \cite{Zhang2018} explored modifications to the decoder of seq2seq models to improve translation results. They used a search engine to retrieve sentences and their translation (referred to as translation pieces) that have high similarity with the source sentence. When similar n-grams from a source document were found in the translation pieces, they rewarded the presence of those n-grams during the decoding process through a scoring mechanism calculating the similarity between source sentence and the source side of the translation pieces. \citet{Zhang2018} reported improvements in translation results up to 6 BLEU points over their seq2seq NMT baseline. In this paper we use the same principle and reward n-grams that are found in the source document during the AMR-to-Text generation process. However we use a simpler approach using a probabilistic language model in the scoring mechanism. 
  
  \section{Guiding NLG for AMR-based summarization}
   
  We first briefly describe the AMR-based summarization method of \citet{Liu2015a} and then our guided NLG approach.
  
  \subsection{AMR-based summarization}
  
  In \citet{Liu2015a}'s work, each of the sentence of the source document was parsed into an AMR graph, and combined into a source graph, $G = (V, E)$ where $v \in V$ and $e \in E$ are the unique concepts and the relations between pairs of concepts. They then extracted a summary graph, $G'$ using the following sub-graph prediction:
  \begin{equation}
    G' = \argmax_{\hat{G} = (\hat{V}, \hat{E})} \sum_{v\in\hat{V}} \boldsymbol{\theta}^\intercal \textbf{f}(v) + \sum_{e\in\hat{E}} \boldsymbol{\psi}^\intercal \textbf{f}(e)
  \end{equation}
  where $\textbf{f}(v)$ and $\textbf{f}(e)$ are the feature representations of node $v$ and edge $e$ respectively. The  final summary produced was a bag of concept words extracted from $G'$. This output we will be replacing with our proposed guided NLG.
  
  \subsection{Unguided NLG from AMR}
    
  Our baseline is a standard (unguided) seq2seq model with attention \cite{Luong2015} which consists of an encoder and a decoder. The encoder computes the hidden representation of the input, $\{z_1, z_2, \ldots, z_k\}$, which is the linearized  summary AMR graph, $G'$ from \citet{Liu2015a}, following \citet{VanNoord2017a}'s preprocessing steps. Following this, the decoder generates the target words, $\{y_1, y_2, \ldots, y_m\}$, using the conditional probability $P_{s2s}(y_j|y_{<j}, z)$,  which is calculated using the equation
    \begin{equation}
      P_{s2s}(y_j|y_{<j}, z) = \text{softmax}(\textbf{W}_s\tilde{\textbf{h}}_t)
    \end{equation}  
  , where the attentional hidden state, $\tilde{\textbf{h}}_t$ is calculated using the equation 
    \begin{equation}
      \tilde{\textbf{h}}_t = \text{tanh}(\textbf{W}_c[\textbf{c}_t; \textbf{h}_t])
    \end{equation}
  ,  where $\textbf{c}_t$ is the source context vector, and $\textbf{h}_t$ is the target RNN hidden state. 
    The source context vector is defined as the weighted average over all the source RNN hidden states, $\bar{\textbf{h}}_s$, given the alignment vector, $\textbf{a}_t$ where $\textbf{a}_t$ is defined as
    \begin{equation}
      \textbf{a}_t(s) = \frac{\text{exp}(\text{score}(\textbf{h}_t, \bar{\textbf{h}}_s))}{\sum_{s'}\text{exp}(\text{score}(\textbf{h}_t, \bar{\textbf{h}}_{s'}))}
    \end{equation}
      
    \subsection{Guided NLG from AMR}
  
    Our goal is to improve the text generated from the summary AMR graph by the probability distribution of the seq2seq model, $P_{s2s}$ using the source document. Since not all sentences in the source document will be used in generating the summary, we prune the source document to a set of $k$ sentences which have the highest similarity with the summary AMR graph. For graph-to-graph similarity comparison, we use the source document AMR parses and calculate the Longest Common Subsequence (LCS) between the linearized AMR parses and the summary AMR graph. We keep the top-$k$ sentences sorted by LCS length. To distinguish this pruned document from the source document, we refer to the former as side information. 
    
    Our aim is is to combine $P_{s2s}$ with the probability distribution estimated using words in the side information, $P_{side}$, in order to score each word given its context during decoding. We estimate $P_{side}$ as the linear interpolation of $2$-gram to $4$-gram probabilities in the form of
    \begin{equation}
      \begin{split}
      P_{side}(x_j|x_{j-3}^{j-1}) &=\lambda_{3} P_{LM}(x_j|x_{j-3}^{j-1})\\
      &+ \lambda_{2} P_{LM}(x_j|x_{j-2}^{j-1}) \\
      &+ \lambda_{1} P_{LM}(x_j|x_{j-1})
      \end{split}
    \end{equation}
   ,  where $x_j$ is a word occurring in side information document, $P_{LM}$ is an $N$-gram LM estimated using Maximum Likelihood:
    \begin{equation}
      P_{LM}(x_j|x^{j-1}_{j-N-1}) = \frac{\text{count}(x_{j-N-1} \ldots x_j)}{\text{count}(x_{j-N-1} \ldots x_{j-1})} 
    \end{equation}
    and $\lambda_i$ is defined as
    \begin{equation}
      \lambda_i = \theta \lambda_{i-1}\,\,\text{where $\theta \in \mathbb{R}$, $\lambda_i > 0$ and $\sum_i \lambda_i = 1$}
    \end{equation}
    where $\theta$ is a hyper-parameter that we tune using the dev dataset during the experiments.
  
    Lastly, we combine the probability distribution of the decoder, $P_{s2s}$ with that provided by the side information, $P_{side}$, as follows:
    \begin{equation}
      \label{eq:score}
      s(y_j|y_{<j}, z) = log\,a 
      + \psi * log (\frac{b}{a} + 1)
    \end{equation}
    where $\psi$ is a hyper-parameter determining the influence of the side information on the decoding process, $a$ is $P_{s2s}(y_j|y_{<j}, z)$ and $b$ is $P_{side}(y_j|y_{j-3}^{j-1}$). $s(y_j|y_{<j}, z)$ is used during beam search replacing $P_{s2s}(y_j|y_{<j}, z)$ for all words that occur in the side information. The intuition behind Eq.~\ref{eq:score} is that we are rewarding word $y_j$ when it appears in similar context in the side information, i.e.\ the source document being summarized.
  
  \section{Experiments}
  We conduct experiments in order to answer the following questions about our proposed approach: (1) Is our baseline model comparable with the state-of-the-art AMR-to-text approaches? (2) Does the guidance from the source document improve the result of AMR-to-Text in the context of summarization? (3) Does the improvement in AMR-to-Text hold when we use the generator for abstractive summarization using AMR? We answer each of these in the following paragraphs.
  
  \paragraph{AMR-to-Text baseline comparison} We compare our baseline model (described in \S3.2) against previous works in AMR-to-text using the data from the recent SemEval-2016 Task 8 \cite[LDC2015E86]{May2016}. Table \ref{table:preprocessing} reports BLEU scores comparing our model against previous works. Here, we see that our model achieves a BLEU score comparable with the state-of-the-art, and thus we argue that it is sufficient to be used in our subsequent experiments with guidance.   
    
    \begin{table}
      \centering
      \small
      \begin{tabular}{l|l}
      Model     & BLEU  \\ \hline
      Our model (unguided NLG) & 21.1 \\ \hline
      NeuralAMR \cite{Konstas2017}    & 22.0  \\ 
      TSP \cite{Song2016a}            & 22.4  \\ 
      TreeToStr \cite{Flanigan2016}    & 23.0    \\ 
      \end{tabular}
      \caption{\label{table:preprocessing}Results for AMR-to-text}
    \end{table}
      
  \paragraph{Guided NLG for AMR-to-Text} In this experiment we apply our guided NLG mechanism described in \S3.3 to our baseline seq2seq model. To isolate the effects of guidance we skip the actual summarization process and proceed to directly generating the summary text from the gold standard summary AMR graphs from the Proxy Report section. To determine the hyper-parameters, we perform a grid search using the dev dataset, where we found the best combination of $\psi$, $\theta$ and $k$ are 0.95, 2.5 and 15 respectively. We have two different settings for this experiment: the oracle and non-oracle settings. In the oracle setting, we directly use the gold standard summary text as the guidance for our model. The intuition is that in this setting, our model knows precisely which words should appear in the summary text, thus providing an  upper bound for the performance of our guided NLG approach. In the non-oracle setting, we use the mechanism described in \S3.3. We also compare them against the baseline (unguided) model from \S3.2. Table ~\ref{table:NLGguidance} reports performance for all models. The difference between the guided and the unguided model is 16.2 points in BLEU and 9.9 points in ROUGE-2, while there is room for improvement as evidenced by the difference between the oracle and non-oracle result.
    
    \begin{table}[h!]
      \centering
      \small
      \begin{tabular}{p{3cm}|l|l|l|l}
      \multirow{2}{*}{Model} & \multirow{2}{*}{BLEU} & \multicolumn{3}{l}{$F_1$ ROUGE} \\ \cline{3-5}
      & & R-1 & R-2 & R-L \\ \hline
      Guided NLG (Oracle) & 61.3  & 79.4 & 63.7 & 76.4      \\
      Guided NLG        & 45.8  & 70.7 & 49.5 & 64.9   \\ \hline
      Unguided NLG        & 29.6  & 68.6 & 39.6 & 61.3 
      \end{tabular}
      \caption{\label{table:NLGguidance}BLEU and ROUGE results for guided and unguided models using test dataset.}
    \end{table}
    
  \paragraph{Guided NLG for full summarization} In this experiment we combine our guided NLG model with \citet{Liu2015a}'s work in order to generate fluent texts from their summary AMR graphs using the hyper-parameters tuned in the previous paragraph. \citet{Liu2015a} used parses from both the manual annotation of the Proxy dataset as well as those obtained using the JAMR parser \cite{Flanigan2014}. Instead of JAMR we use the RIGA parser \cite{Barzdins2016} which had the highest accuracy in the SemEval 2016 Task 8 \cite{May2016}. We compare our result against \citet{Liu2015a}'s bag of words\footnote{We were able to obtain comparable AMR summarization subgraph prediction to their reported results using their published software but not to match their bag-of-word generation results.}, the unguided AMR-to-text model from \S3.2, and a seq2seq summarization model (OpenNMT BRNN)\footnote{We use the OpenNMT-pytorch implementation \url{https://github.com/OpenNMT/OpenNMT-py} and a pre-trained model downloaded from \url{http://opennmt.net/OpenNMT-py/Summarization.html} which has higher result than \citet{See2017}'s summarizer.}\footnote{The pre-trained model generates multiple sentences summary, but we use only the first sentence summary for evaluation in accordance with the AMR dataset.} which summarizes directly from the source document to summary sentence without using AMR as an interlingua and is trained on CNN/DM corpus \cite{Hermann2015} using the same settings as \citet{See2017}. 
    
    \begin{table}[h!]
      \centering
      \small
      \begin{tabular}{p{1cm}|p{2.2cm}|S[table-format=3.1]|S[table-format=3.1]|S[table-format=3.1]}
       AMR & \multirow{2}{*}{NLG Model} & \multicolumn{3}{l}{$F_1$ ROUGE} \\ \cline{3-5}
        parses & & {R-1} & {R-2} & {R-L} \\ \hline
       \multirow{3}{*}{Gold}& Guided & 40.4  & 20.3 & 31.4  \\
           & Unguided & 38.9  & 12.9 & 27.0 \\
              & \citet{Liu2015a} & 39.6 & 6.2 & 22.1 \\ \hline
      \multirow{3}{*}{RIGA}& Guided & 42.3 & 21.2 & 33.6  \\
      & Unguided & 37.8 & 10.7 & 26.9 \\
          & \citet{Liu2015a} & 40.9 & 5.5 & 21.4 \\ \hline 
           Directly from Text & OpenNMT BRNN 2 layer, emb 256, hidden 1024 & 36.1 & 19.2 & 31.1
      \end{tabular}
      \caption{\label{table:summarizationguidance} The $F_1$ ROUGE scores for guided, unguided, \citet{Liu2015a} (BoW) results in Gold and RIGA parses, and seq2seq summarization. All models are run using test dataset.}
    \end{table}
  
  In Table~\ref{table:summarizationguidance}, we can see that our approach results in improvements over both the unguided AMR-to-text and the standard seq2seq summarization. One interesting note is that using the RIGA parses result in higher ROUGE scores than the gold parses for the guided model in our experiment. This phenomenon was also observed in \citet{Liu2015a}'s experiment where the summary graphs extracted from automatic parses had higher accuracy than those extracted from manual parses. We hypothesize this can be attributed to how the AMR dataset is annotated as there might be discrepancies in different annotator's choices of AMR concepts and relations for sentences with similar wording. In contrast, the AMR parsers introduce errors, but they are consistent in their choices of AMR concepts and relations. The discrepancies in the manual annotation could have impacted the performance of the AMR summarizer that we use more negatively than the noise introduced due to the AMR parsing errors.
  \begin{table}[h!]
    \centering
    \small
    \begin{tabular}{l|p{5cm}}
    NLG Model & Generated Summary \\ \hline
    Gold & on 8 august 2008 russia conducted \textbf{airstrikes} on \textbf{georgian} targets .\\ \hline
    Guided &  on 8 august 2008 russia conducted \textbf{airstrikes} on \textbf{georgian} separatist targets .\\
    Unguided & on 8 august 2008 russia conducted \textbf{a softening} of the \textbf{georgia 's} separatist target . \\ \hline
    Seq2seq & the russian laboratory complex is a 90 - building campus and served as the location for russia 's secret biological weapons program in the soviet era of a moscow regional depository threaten moscow . \\
    \end{tabular}
    \caption{\label{table:NLGSummary} Result summaries of guided, unguided and seq2seq models compared with gold summary.}
  \end{table}
  
  In Table \ref{table:NLGSummary}, we show sample summaries from the different models, where we can see that our guided model improves the unguided model by correcting a wrong word (\textit{a softening}) into a correct one (\textit{airstrikes}) and introducing a better suited word from the source document (\textit{georgian} instead of \textit{georgia 's}). 
  
  \begin{table}[h!]
    \centering
    \small
    \begin{tabular}{l|l}
    NLG Model & Fluency \\ \hline
    Guided & 2.66 \\
    Unguided & 2.16 \\
    \end{tabular}
    \caption{\label{table:NLGHuman} Fluency scores on test dataset.}
  \end{table}

  We also evaluated manually by asking human evaluators to judge sentences' fluency (grammatical and naturalness) on a scale of 1 (worst) to 6 (best) for the guided and unguided model (see Table \ref{table:NLGHuman}). While the manual evaluation shows improvement over the unguided model, on the other hand, grammatical mistakes and redundant repetition in the generated text are still major problems (see Table \ref{table:NLGMistakes}) in our AMR generation.
  
  \begin{table}[h!]
    \centering
    \small
    \begin{tabular}{p{3.8cm}|l}
    Guided NLG Model & Problems \\ \hline
    the soldiers were injured when \textbf{a attempt} to defuse the bombs .& grammatical mistake \\
    \textbf{on 20 october 2002} the state - run radio nepal reported \textbf{on 20 october 2002} that at the evening \textbf{- run radio nepal reported on 20 october 2002} that the guerrillas were killed \textbf{and killed} . & redundant repetition \\
    \end{tabular}
    \caption{\label{table:NLGMistakes} Problems in guided model's summaries.}
  \end{table}

  \section{Conclusion and Future Works}
  
  In this paper we proposed a guided NLG approach that substantially improves the output of AMR-based summarization. Our approach uses a simple guiding process based on a probabilistic language model. In future work we aim to improve summarization performance by jointly training the guiding process with the AMR-based summarization process. 
  
  \section*{Acknowledgments}
  We are thankful for Gerasimos Lampouras for his help with the manual evaluation process and all volunteers who participated in it. We would also like to thank the Indonesian government that has sponsored the first author's studies through the Indonesia Endowment Fund for Education (LPDP). The second author is supported by the EU H2020 SUMMA project (grant agreement number 688139) and the EPSRC grant eNeMILP (EP/R021643/1).
  
  \bibliography{emnlp2018}
  \bibliographystyle{acl_natbib_nourl}
  
  \end{document}